# Is Machine Learning Speaking my Language? A Critical Look at the NLP-Pipeline Across 8 Human Languages

Esma Wali [1] Yan Chen[1] Christopher M Mahoney[1] Thomas G Middleton[1] Marzieh Babaeianjelodar[1] Mariama Njie[2] Jeanna Neefe Matthews[1]

## Abstract

Natural Language Processing (NLP) is increasingly used as a key ingredient in critical decision-making systems such as resume parsers used in sorting a list of job candidates. NLP systems often ingest large corpora of human text, attempting to learn from past human behavior and decisions in order to produce systems that will make recommendations about our future world. Over 7000 human languages are being spoken today and the typical NLP pipeline underrepresents speakers of most of them while amplifying the voices of speakers of other languages. In this paper, a team including speakers of 8 languages - English, Chinese, Urdu, Farsi, Arabic, French, Spanish, and Wolof - takes a critical look at the typical NLP pipeline and how even when a language is technically supported, substantial caveats remain to prevent full participation. Despite huge and admirable investments in multilingual support in many tools and resources, we are still making NLP-guided decisions that systematically and dramatically underrepresent the voices of much of the world.

## 1. Introduction

Corpora of human language are regularly fed into machine learning systems as a key way to learn about the world. Systems are taught to identify spam email, suggest medical articles or diagnoses related to a patient's symptoms, sort resumes based on relevance for a given position and many other tasks that form key components of critical decision making systems in areas such as criminal justice, credit, hiring, housing, allocation of public resources, medical decisions and more. Much like facial recognition systems are often trained to represent white men more than black women (Buolamwini, J., 2018), machine learning systems are often trained to represent human expression in languages such as English more than in languages such as Urdu or Wolof. In this paper, a team including speakers of 8 languages (native speakers of 5) ask what it would take to see all the languages we speak fully included in NLP-based research?

A typical NLP-pipeline includes steps such as gathering corpora, processing them into text format, identifying key language elements, training models, and then using these models to answer predictive questions. For some languages,

[1] Department of Computer Science, Clarkson University, Potsdam, NY, USA

[2] Department of Computer Science, Iona College, NY, USA

[1] Esma Wali <walie@clarkson.edu> Yan Chen <cheny3@clarkson.edu> Christopher M Mahoney <mahonec@clarkson.edu> Thomas G Middleton <tmiddlet@clarkson.edu> Marzieh Babaeianjelodar <babaeim@clarkson.edu> Jeanna Neefe Matthews<jmatthew@clarkson.edu> [2] Mariama Njie<mnjie1@gaels.iona.edu>

Is Machine Learning Speaking my Language? A Critical Look at NLP-Pipeline Across 8 Human Languagesthere are well-developed resources available throughout the stages of this pipeline. For some languages, pre-trained models even exist allowing research or development teams to jump right to the last step.

Pre-training from scratch using the large corpora necessary for meaningful NLP-results is expensive (i.e. days on a dozen CPUs). When a team can download a pre-trained model, they avoid this substantial overhead. Fine-tuning is much less expensive (i.e. hours on a single CPU). This makes NLP-based results accessible to a wider range of people, but only if such a pre-trained model is available for their language. When these easy to use pre-trained models exist for only a few languages, it further exacerbates the disparity in representation and participation.

It is increasingly common to use these pre-trained models without a clear evaluation/understanding of the ingredients used to build them. For example, a recent paper from Babaeianjelodar et al. demonstrated surprising differences in gender bias when starting with a pre-trained BERT model based on a "representative" Wikipedia and Book Corpus in English and then fine-tuning with various smaller corpora including the GLUE benchmarks and several corpora of hate speech (Babaeianjelodar et al., 2020). Bolukbasi et al's pioneering work in quantifying gender bias used Word2Vec on a corpus of Google News in English (Bolukabsi et al., 2016).

The degree to which some languages are under-represented in commonly used text-based corpora is well-recognized, but the ways in which this effect is magnified throughout the NLP-tool chain and the ways in which even tools that support a language come with substantial caveats are less discussed. Despite huge and admirable investments in multilingual support in project like Wikipedia (Wikipedia, 2020C), BERT (Devlin et al., 2018), Word2Vec (Mikolov et al., 2013), Wikipedia2Vec (Yamada et al., 2016; Ousia, 2016), Natural Language Toolkit (NLTK) (NLTK, 2005), MultiNLI (Williams et al., 2020), and many more, we are still making NLP-guided decisions that systematically and dramatically underrepresent the voices of much of the world. As speakers of 8 languages who have recently examined the modern NLP toolchain, we highlight the difficulties that speakers of many languages face in having their thoughts and expressions included in the NLP-derived conclusions that are being used to direct the future for all of us.

We see connections between this work and other participatory ML work to analyze the amplifiers of systemic injustice in decision systems and give increased voice to communities affected by ML systems. Halfaker et al. discusses lowering technical barriers to allow more participation in ML classifiers for Wikipedia edit moderation by using ORES (Halfaker & Geiger, 2019). Brown et al. discusses the importance of improving the accountability of algorithms by incorporating feedback from parties who are impacted by the decisions being made, particularly families receiving public assistance (Brown et al., 2019). Katell et al. similarly speaks to the importance of non-technical feedback into algorithms to improve accountability (Katell et al., 2020). Patton et al. addresses the use of African American vernacular English on social media as it pertains to classifying for use by social workers in gang interventional services (Patton et al., 2020).

A recent work in this area is carried out by (Joshi et al., 2020) where they looked at the relation between languages, resources and NLP conferences to shed light on the trajectory different languages followed over time. They illustrated how low resourced languages get attention by tracking the number of authors writing about different classes of languages in a range of NLP conferences. They found disparities among the languages by showing that many authors and most of the older language conferences have been focusing on already resource rich languages while inclusion of low resourced languages has been minimal. It is worth noting that some of the newer conferences and communities like International Conference on Computational Linguistics (CL) and International Conference on Language, Resource and Evaluation (LREC) have been more language inclusive over the period of time.

Ponti et al also promote a large-scale typology that provides guidance for multilingual (NLP), particularly for languages that suffer from the lack of human-labeled resources. They believe existing NLP is still largely limited to a handful of resource-rich languages and they advocate for a new approach that adapts the broad and discrete nature of typological categories to the contextual and continuous nature of machine learning algorithms used in contemporary NLP (Ponti et al., 2018).

Bender (2011) holds a critical lens to language independence claims of computational linguistics and NLP systems while arguing the importance of including linguistic typology knowledge in NLP system development to improve language independence.

## 2. NLP-Tools

The vast majority of NLP-Tools are first developed for English and even when support for other languages is added it often lags behind in robustness, accuracy and efficiency. The evolution of BERT (Bidirectional Encoder Representations from Transformers) from Google offers a good example (Delvin J., 2019).

The original BERT models released in 2018 were English-only, but soon after a Chinese model and a Multilingual model were released. However, single language models are acknowledged to have advantages over the multilingual model. For example, the BERT Github page (Devlin, J., 2019) said that while the multilingual model supports English

**Is Machine Learning Speaking my Language? A Critical Look at NLP-Pipeline Across 8 Human Languages**and Chinese that the Chinese specific model would likely produce better results for fine-tuning with Chinese-only data.

Similarly, when advances are made, they are often available only in English. For example, in March 2020, 24 smaller/condensed BERT models were released. These condensed models were intended to help teams with restricted computational resources, but all 24 were English only.

The Multilingual BERT model was designed and tested on a subset of languages. Specifically, it was evaluated using the XNLI dataset (Conneau et al., 2018), which is a version of MultiNLI where the dev and test sets have been translated (by humans) into 15 languages. For these 15 languages, they show patterns of lower accuracy for some languages and also show that the individual language models for English and Chinese give ~3% advantage for those languages. It is an advantage to be included in the list of 15 languages for which the evaluation was completed.

Lack of representation at each stage of the pipeline adds to lack of representation in later stages of the pipeline. For example, the Multilingual BERT GitHub page (Devlin, J., 2019) says that the data used for the Multilingual model was the top 100 languages with the largest Wikipedias. The entire Wikipedia dump for each language (excluding user and talk pages) was taken as the training data for each language. As we will discuss later in this paper, there are substantial differences in the size and quality of the Wikipedia corpus for different languages even when adjusting for the number of speakers. This is despite a large and admirable investment in the development of multilingual Wikipedia corpora.

Moving beyond the example of BERT to include other NLP tools, in Table 1 we list the 8 languages we examined (sorted by the number of speakers of that language worldwide) and indicate whether the common NLP tools BERT, Word2Vec, NLTK and Wikipedia2Vec support them.

Word2Vec is a collection of two-layered shallow neural networks used to produce word embeddings (Mikolov et al., 2013). Bolukbasi et al. (2016) used English word embeddings to calculate gender bias in Google News Corpus. Its usage ranges from sentiment analysis and classification (Zhang et al., 2015), text classification (Lilleberg et al., 2015), named entity recognition (Sienčnik, 2015), computing word similarity (TextMiner, 2017), computation of gender bias (Bolukbasi et al., 2016) and many more. The TextMiner (2017) trained the Word2Vec model on Wikipedia by Gensim to compute word similarity with support for over 80 languages.

NLTK has support provided for dozens of corpora and trained models (NLTK, 2005). However, if we analyze this support regarding various functions like word tokenizing, NLTK stopwords, and sentiment analysis (Lo et al., 2005; Stackoverflow, 2013), we find limited support for many languages. We see some efforts (Ashraf, 2018) to develop NLTK for some Indian languages and Urdu is also included in this.

**Table 1: Support for 8 languages in several common NLP-tools**

| Language | BERT | Word2Vec | NLTK | Wikipedia2Vec |
|---|---|---|---|---|
| Chinese | ✓ | ✓ | ✓ | ✓ |
| English | ✓ | ✓ | ✓ | ✓ |
| Spanish | ✓ | ✓ | ✓ | ✓ |
| Arabic | ✓ | ✓ | ✓ | ✓ |
| French | ✓ | ✓ | ✓ | ✓ |
| Farsi | ✓ | ✓ | ✗ | ✗ |
| Urdu | ✓ | ✗ | ✓ | ✗ |
| Wolof | ✗ | ✗ | ✗ | ✗ |

However, the framework was not suggested to use (Victoroff, 2017) regarding support provided for a broad array of human languages. For our set of 8 languages, NLTK support is not available for Farsi and Wolof.

Wikipedia2Vec, a Python-based open-source, optimized tool for learning the embeddings of words and entities from Wikipedia, provides pre-trained embeddings for 12 languages in binary and text format (Yamada et al., 2016; Ousia, 2016).

All four tools listed in Table 1 technically support the 5 most spoken languages in our list, but support for Farsi, Urdu and Wolof is spottier.

It is also important to recognize that a check mark indicating support does not tell the whole story. Notice that we have listed BERT as supporting all of the languages except Wolof. However, as we have discussed above, there are some serious caveats to that support (lack of downloadable pre-trained models, lack of condensed pre-trained models, lower accuracy, lack of testing, etc.) Exceptions of this kind are common across all NLP tools even when they do technically support a given language.

## 3. Wikipedia and Book Corpora

Many pre-trained models available for BERT were constructed using a data set consisting of English Wikipedia with 2500 million words and a BookCorpus with 800 million English words. For each of the 8 languages that we speak, we consider how we could construct a similar dataset based on Wikipedia and a set of books written originally in that language.

We begin by discussing the Wikipedia corpora available for each of our 8 languages. Of the over 7000 languages spoken today only approximately 300 have a Wikipedia corpus, including all 8 of the languages on which we are focusing (Eberhard et al., 2020). For these languages, Wikipedia itself provides a rich set of metrics on the differing characteristics of the corpora for each language represented (Wikimedia, 2015). For example, they compare the number of speakers of a language to the number of articles in the Wikipedia Corpora for that language with the metrics Articles/1000 Speakers. The many contributors to Wikipedia have made a large and admirable investment in assembling a large multilingual dataset. In fact, Wikipedia is the largest multilingual online knowledge repository (Wikipedia, 2020C) and as we discussed earlier, it has played a key role in the development of multilingual NLP tools. However, still, the vast majority of languages spoken today are not represented.

In Table 2, we list the 8 languages considered in this paper. It is interesting to note that the number of speakers of each language does not track evenly with the number of articles. French has the highest ratio of Articles/1000 Speakers at 29.7 and Wolof the lowest at 0.14. English has the largest number of articles, even including languages not in our list, but its ratio of articles to speakers is lower than some languages.

Certainly, not all speakers of a language have equal access to contributing to Wikipedia. In the case of Chinese, Chinese speakers in mainland China have little access to Wikipedia because it is banned by Chinese government (Siegel, R, 2019). Thus, Chinese articles in Wikipedia are more likely to have been contributed by the 40 million Chinese speakers in Taiwan, Hong Kong, Singapore and overseas (Su, 2019). In other cases, the percentage of speakers with access to Wikipedia may vary for other reasons such as access to computing devices and Internet access.

Interestingly, Wikimedia reports 5,378,533 articles in a corpus for Cebuano, a language spoken in the Philippines (Wikimedia, 2020). This is the second largest number of articles after English and with an approximate 16 million number of speakers that would be 340.2 Articles/1000 speakers. Volapük (an international auxiliary language invented by Johann Martin Schleyer, a German Priest, in 1879 and 1880) has the largest ratio of 623,560 Articles/1000 speakers.

**Table 2: Comparing the number of speakers of a language to the size of the Wikipedia Corpora for that language.** For the number of articles and estimates of the number of speakers of Chinese, English, Spanish, Arabic, French, Urdu and Farsi, we used Wikipedia's article on the "List of Wikipedias by speakers per article" (Wikimedia, 2015). For Wolof, we obtained the number of speakers from Wikipedia's article on Wolof (Wikipedia, 2020A), and the number of articles from the statistics of the Wolof corpora itself (Wikipedia, 2020B).

| Language | Number of Speakers | Number of Articles (Wikipedia) | Articles/ 1000 Speakers |
|---|---|---|---|
| Chinese | 1197 million | 1,124,594 | 0.94 |
| English | 505 million | 6,102,188 | 12.08 |
| Spanish | 470 million | 1,605,891 | 3.42 |
| Arabic | 315 million | 1,048,391 | 3.33 |
| French | 75 million | 2,227,687 | 29.70 |
| Farsi | 72 million | 732,106 | 10.17 |
| Urdu | 64 million | 155,298 | 2.43 |
| Wolof | 10 million | 1393 | 0.14 |

In addition to the difference in the number of articles and Articles/1000 speakers, Wikipedia corpora for different language also vary widely along many other dimensions including the total size of corpora in MB, total pages, percentage of articles that are simply stub articles with no content, number of edits, number of admins working in that language, total number of users and total number of active users. Thus, even though there are Wikipedia corpora for all 8 languages they differ substantially in size and quality. Once again, a checkmark saying that a Wikipedia corpus exists hides many caveats to full representation and participation.

We also encountered hurdles in processing different Wikipedia corpora. For example, additional processing is required before Wikipedia documents in an XML format are ready to be ingested into models such as Word2Vec or BERT (e.g. removal of the XML markup tags). We downloaded the Chinese, English, Spanish, Arabic, French and Farsi corpora from Linguatools and they provided a script (xml2txt.pl) for post-processing (Linguatools, 2018). However, we found differences in the number of errors when using this script with different languages. We saw no errors across the over 5 million English articles processed, but for Farsi, we saw an error rate of 0.13% and for Chinese, we saw an error rate of 0.02%.

Further, this script was not even applicable to the Urdu and Wolof corpora because they were downloaded in a different format from Wikimedia dumps (Wikimedia Dumps, 2020) when they were not available in the set provided by Linguatools. For Urdu and Wolof, the script would need to be modified or a new script written, an additional hurdle.

Moving beyond these Wikipedia corpora, we also assembled a list of classic books in each language (e.g "Cien Años de Soledad" in Spanish and "Le Petit Prince" in French) and investigated the hurdles we would face in adding these books into our corpora for training. For some languages, we found no problem in finding text versions for download. However, for other languages, we encountered a variety of hurdles.

For Arabic and Urdu, we found many texts available as scanned images rather than text format. With wide ranges of accuracy rates for OCR of 70 to 98% (Holley, 2009), OCR to convert scanned images to text can be problematic in any language (Tiercelin, 2009). Still, we noticed more errors in some languages.

In Chinese, we observed that OCR incorrectly added space every time a new line began and we observed many wrong characters (e.g 猫(correct) -> 犾(incorrect)). Some OCR tools are developed for specific languages and work quite efficiently (Sakhr, 1995). Interestingly, in some languages like Chinese, we found it easier to get freely downloadable full text versions than in other languages. This may reflect societal/cultural differences in approaches to intellectual property/copyright restrictions.

Interestingly, the Wolof language does not have a written character set of its own and therefore text versions of Wolof are represented using English, French and Arabic characters. The same Wolof passage could be transcribed differently in the different character sets. For example the surname Njie (a common surname in the Wolof tribe) is written as such in English but written as N'Diaye in French. Another factor for all 8 languages but especially Wolof is the processing of audio corpora rather than image or text representations of written text (Gauthier et al., 2016).

## 4. Conclusion

Despite huge and admirable investments in multilingual support in projects like Wikipedia, BERT, Word2Vec, Wikipedia2Vec, and NLTK, we are still making NLP-guided decisions that systematically and dramatically underrepresent the voices of much of the world. Even when tools technically do support a given language, there are often substantial caveats such as higher rates of error and lack of testing that prevent full participation/representation. We document how lack of representation in the early stages of the NLP pipeline (e.g. representation in Wikipedia) is further magnified throughout the NLP-tool chain, culminating in reliance on easy-to-use pre-trained models that effectively prevents all but the most highly resourced teams from including diverse voices. As speakers of 8 languages, we highlight the difficulties that speakers of many languages still face in having their thoughts and expressions fully included in the NLP-derived conclusions that are being used to direct the future for all of us.

## 5. Acknowledgements

We gratefully acknowledge the input and assistance of our wider research team including Hunter Bashaw, Josh Gordan, Izzi Grasso, Stephen Lorenz, Abigail Matthews, Graham Northup and Cameron Weinfurt. We appreciate the insight of Clarkson faculty Golshan Madraki and Yu Liu. Support from the Clarkson Open Source Institute was essential to this work.

Thanks to baby Anna who has been patiently attending our weekly meetings even though she won't come to Clarkson until 2038.